%% file: main.tex
\documentclass[manuscript]{acmart}
\AtBeginDocument{%
  }
\usepackage{amsmath}
\usepackage{algorithm}
\usepackage{algpseudocode}
\usepackage{multirow}
\usepackage{makecell}
\usepackage{booktabs}
\usepackage{array} 
\usepackage{graphicx}
\usepackage{rotating}
\usepackage{listings}
\usepackage{xcolor}
\usepackage{afterpage}
\usepackage{placeins}
\usepackage[caption=false,font=footnotesize,labelfont=rm,textfont=rm]{subfig}
\DeclareSubrefFormat{parens}{#1(#2)}
\setcopyright{acmlicensed}
\copyrightyear{2018}
\acmYear{2018}
\acmDOI{XXXXXXX.XXXXXXX}
\acmConference[Conference acronym 'XX]{Make sure to enter the correct
  conference title from your rights confirmation email}{June 03--05,
  2018}{Woodstock, NY}
\acmISBN{978-1-4503-XXXX-X/2018/06}

\begin{document}

\title{TableMind++: An Uncertainty-Aware Programmatic Agent for Tool-Augmented Table Reasoning}

\author{Mingyue Cheng}
\author{Shuo Yu}
\author{Chuang Jiang}
\author{Xiaoyu Tao}
\author{Qingyang Mao}
\author{Jie Ouyang}
\author{Qi Liu}
\author{Enhong Chen}
\affiliation{%
  \institution{State Key Laboratory of Cognitive Intelligence, University of Science and Technology of China}
  \city{Hefei}
  \country{China}}
\email{mycheng@ustc.edu.cn}
\email{yu12345@mail.ustc.edu.cn}
\email{jiangchuang@mail.ustc.edu.cn}
\email{txytiny@mail.ustc.edu.cn}
\email{ouyang_jie@mail.ustc.edu.cn}
\email{qiliuql@ustc.edu.cn}
\email{cheneh@ustc.edu.cn}

\renewcommand{\shortauthors}{M. Cheng et al.}

\begin{abstract}
\input{0-Abstract}
\end{abstract}

\ccsdesc[500]{Computing methodologies~Intelligent agents}

\keywords{Table reasoning, Programmatic Agent, LLMs}


\maketitle

\input{1-Introduction}
\input{2-RelatedWork}
\input{3-Preliminaries}
\input{4-SystemDesign}

\input{4.5-SystemDesign}

\input{5-Experiments}

\input{6-Conclusion}

\bibliographystyle{ACM-Reference-Format}
\bibliography{main}

\appendix

\end{document}

%% file: 0-Abstract.tex
Table reasoning requires models to jointly perform semantic understanding and precise numerical operations. Most existing methods rely on a single-turn reasoning paradigm over tables which suffers from context overflow and weak numerical sensitivity. To address these limitations, we previously proposed TableMind as a tuning-based autonomous programmatic agent that simulates human-like interaction within a lightweight large language model (LLM). TableMind internalizes planning, action, and reflection through a two-stage training strategy involving supervised fine-tuning (SFT) on filtered high-quality data and reinforcement learning (RL) via a multi-perspective reward and the Rank-Aware Policy Optimization (RAPO) algorithm. While TableMind establishes a solid foundation for programmatic agents, the inherent stochasticity of LLMs remains a critical challenge that leads to hallucinations. In this paper, we extend this foundation to TableMind++ by introducing a novel uncertainty-aware inference framework to mitigate hallucinations. Specifically, we propose memory-guided plan pruning to retrieve historical trajectories for validating and filtering out logically flawed plans to address epistemic uncertainty. To ensure execution precision, we introduce confidence-based action refinement which monitors token-level probabilities to detect and self-correct syntactic noise for aleatoric uncertainty mitigation. Finally, we employ dual-weighted trajectory aggregation to synthesize a robust consensus from multiple reasoning paths. Extensive experiments on diverse benchmarks demonstrate that TableMind++ consistently outperforms previous baselines and proprietary models to validate the effectiveness of integrating autonomous training with uncertainty quantification.
Our code is available\footnote{\url{https://github.com/fishsure/TableMind-PP}}.

%% file: 1-Introduction.tex
\begin{figure*}[htbp]
    \centering
    \includegraphics[width=1\linewidth]{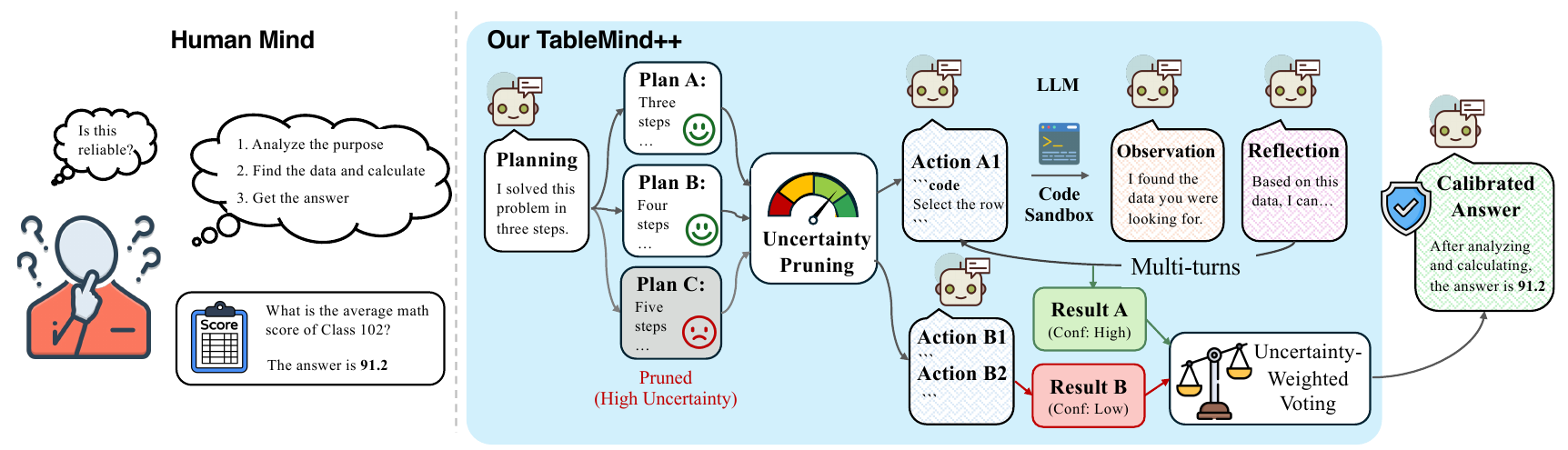}
    \caption{TableMind++ mimics the human chain of thought by using a multi-turn plan-action-reflect loop to solve table tasks. This foundational loop is augmented with uncertainty-aware guardrails including planning pruning  and weighted aggregation during inference to filter errors and guarantee reliable execution.}
    \label{fig:overView}
\end{figure*}

\section{Introduction}
\label{sec:intro}
Tables are one of the most fundamental forms of structured data, serving as a ubiquitous medium for knowledge storage and exchange. They are widely utilized to organize high-value information across diverse domains, including experimental logs in scientific research~\cite{hollmann2025accurate}, patient records in healthcare management~\cite{pampari2018emrqa}, and financial statements in business intelligence~\cite{chen2022convfinqa}. Consequently, effectively mining this vast amount of structured knowledge has become a practically important research problem~\cite{ye2025closer}. Unlike unstructured text processing, this task presents unique challenges as it requires models to jointly handle comprehensive semantic understanding of schema and cell content, while simultaneously performing precise numerical operations such as arithmetic calculation and logical comparison.

Despite their effectiveness, most existing methods still adopt a single-turn reasoning paradigm by processing flattened tables in a single forward pass, which leads to fundamental limitations~\cite{cheng2025survey}. First, this paradigm exhibits significant structural rigidity. By flattening tables into textual inputs, LLM-based approaches~\cite{tao2025values} are vulnerable to context overflow and inherently insensitive to continuous numerical values. Treating continuous data as plain tokens hinders precise arithmetic and logical operations, frequently leading to unstable reasoning and unacceptable calculation errors~\cite{zhou2025benchmarking}. Second, existing methods suffer from intrinsic and severe inference unreliability. Most models rely on the black-box capabilities of LLMs without explicit mechanisms for tool-use, execution monitoring, and reflection~\cite{cheng2025agentr1trainingpowerfulllm}. Critically, the inherent uncertainty of LLMs involving both epistemic ambiguity and aleatoric noise makes tasks demanding rigorous reasoning particularly challenging. Without robust uncertainty quantification , these systems are highly susceptible to probabilistic errors and hallucinations~\cite{yadkori2024believe,kapoor2024large}. Overall, these limitations reveal a significant mismatch between simplistic single-turn generation and the reliable multi-turn interaction required for complex table reasoning.

These analyses motivate a closer examination of how humans cognitively perform table reasoning, as illustrated in the left part of Figure~\ref{fig:overView}. Given a query, humans typically adopt a systematic multi-stage process to derive answers from complex tables. Specifically, they analyze the query and decompose it into sub-tasks, locating relevant rows and columns, performing explicit numerical operations, and checking intermediate results, and finally deriving a validated answer.  A key characteristic of this process is that reasoning proceeds via explicit actions, such as calculations or numerical operations, while intermediate results are repeatedly verified~\cite{chen2019tabfact}.  Crucially, throughout this workflow, humans assess their confidence levels to judge the reliability of their reasoning steps. This observation suggests that reliable table reasoning requires explicit planning, action, and reflection, alongside rigorous uncertainty quantification to mitigate hallucinations, rather than a simplistic single-turn generation over structured tables~\cite{cheng2025survey}.

Motivated by the human-like cognitive schema of the multi-turn interaction, a direction is to leverage LLM-driven agents~\cite{huang2024understanding} to explicitly model such processes, as shown in the right part of Figure~\ref{fig:overView}. In practice, LLMs are coordinated through predefined pipelines to perform such processes, with the most straightforward implementation being training-free workflow-based agents~\cite{yao2023react, wang2024survey}. While effective for numerical reasoning, it comes with notable limitations~\cite{cheng2024towards, zhai2024large}. Workflow-based agents typically rely on multiple LLM calls and external execution environments, resulting in computational overhead. Moreover, the need to expose tabular data to external models raises practical concerns regarding data privacy and security, especially in sensitive real-world applications. Given these limitations, we aim for a more meaningful and challenging goal: instead of relying on heavy workflows, we seek to train a lightweight table agent that intrinsically acquires human-like reasoning capabilities. Such an agent is expected to internalize planning, action, and reflection capabilities within an LLM, enabling efficient, privacy-preserving, and accurate table reasoning. Furthermore, by constructing a memory module and integrating reflection mechanisms during inference, the agent effectively mitigates hallucinations and related issues stemming from the inherent uncertainty of LLMs.

To achieve the goal of internalizing human-like reasoning, our preliminary work introduced TableMind, a lightweight autonomous agent established through a principled two-stage training strategy. In the supervised fine-tuning (SFT) stage, we utilized high-quality reasoning trajectories to bootstrap basic capabilities, including problem decomposition and code generation. Building upon this, we further introduced a reinforcement learning (RL) stage employing Rank-Aware Policy Optimization (RAPO). By identifying misaligned trajectories and amplifying learning signals through multi-view rewards, TableMind successfully learned to internalize a multi-turn cognitive policy within a lightweight LLM, enabling autonomous planning and tool-use without relying on heavy external workflows.

However, even a well-optimized policy remains susceptible to the inherent stochasticity of generative models, which poses significant risks in high-precision tasks. For the sake of mitigating the hallucinations and potential errors stemming from this inherent stochasticity, we further propose TableMind++, which equips the foundational agent with an advanced uncertainty-aware inference framework. Specifically, we introduce memory-guided plan pruning to reduce epistemic uncertainty by validating high-level plans against a dual-memory of historical successes and failures, thereby ensuring that the reasoning logic remains within a verifiable search space. Simultaneously, we employ confidence-based action refinement to manage aleatoric uncertainty, filtering out syntactic noise and logic inconsistencies in code generation through real-time token-level confidence monitoring. Finally, to synthesize a robust consensus, we implement dual-weighted trajectory aggregation, which calibrates the final answer based on both structural validity and execution certainty, effectively suppressing occasional generation outliers. Extensive experiments on diverse public benchmarks demonstrate that TableMind++ consistently outperforms previous baselines, validating the effectiveness of our holistic approach that combines autonomous training with uncertainty-aware inference.

Our main contributions are summarized as follows:

\begin{itemize}
    \item For the sake of mitigating the hallucinations and inconsistency caused by the inherent stochasticity of the base model, we propose TableMind++, an uncertainty-aware autonomous agent that synergizes a robustly trained cognitive policy with a rigorous inference framework to achieve reliable multi-turn table reasoning.
    \item We introduce a two-stage training strategy combining supervised fine-tuning and reinforcement learning with our proposed Rank-Aware Policy Optimization (RAPO) to establish the human-like reasoning capabilities.
    \item We design a dynamic inference framework that serves as a reliability guardrail. By integrating memory-guided plan pruning and confidence-based action refinement, we explicitly mitigate the inherent epistemic and aleatoric uncertainties of LLMs, ensuring trustworthy execution.
    \item We conduct extensive experiments on multiple benchmarks, demonstrating that TableMind++ achieves state-of-the-art performance and consistently outperforms competitive baselines.
\end{itemize}

%% file: 2-RelatedWork.tex
\section{Related Work}

In this section, we review the extensive literature relevant to our work, structured into three primary dimensions:  advancements in LLM-based table reasoning,  the development of autonomous agents and tool learning, and methodologies for uncertainty quantification in generative models.

\subsection{Table Reasoning}

By presenting data in a structured format, tables serve as a powerful means for organizing, storing, and analyzing information. Early approaches primarily focused on learning robust table representations through specialized pre-training tasks, enabling models to grasp the inherent structural constraints of tabular data. Inspired by the success of Masked Language Modeling in BERT~\cite{koroteev2021bert}, a prominent line of work involves masking parts of the table to force the model to recover missing information based on context. For instance, TaPas~\cite{herzig2020tapas} requires the model to reconstruct masked cells, thereby learning the correlation between headers and cell values. Extending this granularity, Pasta~\cite{gu2022pasta} and TUTA~\cite{wang2021tuta} propose masking entire columns or distinct table segments, enabling the model to capture broader row-column dependencies and hierarchical structures. Distinct from these reconstruction-based methods, TAPEX~\cite{liu2021tapex} introduces an execution-centric paradigm. By pre-training an encoder-decoder model on a large-scale synthetic SQL dataset, TAPEX mimics a SQL executor. This approach bridges the gap between natural language and symbolic execution, granting the model a deeper, functional understanding of tabular logic before fine-tuning on downstream tasks.

With the emergence of LLMs, the research focus has shifted from representation learning toward developing general-purpose systems capable of complex reasoning and autonomous execution. Models such as TableLLaMA~\cite{zhang2023tablellama}, TableLLM~\cite{zhang2024tablellm}, and the TableGPT series~\cite{zha2023tablegpt,su2024tablegpt2} adapt general LLMs to structured data through extensive instruction tuning. By combining language reasoning with table parsing, these models achieve strong zero-shot performance without task-specific retraining. To further handle complex multi-step queries, recent research emphasizes workflow-driven or agentic methods. Chain-of-Table~\cite{wang2024chain} follows a dynamic data-flow paradigm, iteratively updating the table content after each reasoning step to simplify the context. Meanwhile, PoTable~\cite{mao2024potable} adopts a plan-then-execute strategy, decoupling high-level planning from low-level execution via external tools (e.g., Python), thus enhancing calculation precision. Most recently, reinforcement learning (RL) has been introduced to optimize the reasoning process itself. As seen in DeepSeek-R1~\cite{guo2025deepseek,luo2025time,wang2025can} and structured data extensions such as Table-R1~\cite{yang2025table} and STaR~\cite{zhang2025star}, RL frameworks incentivize models to self-correct and evolve their reasoning strategies. This evolution~\cite{wang2024tabletime} underscores the growing significance of combining generative capabilities with rigorous logic for robust table reasoning.

\subsection{LLM-based Agent}

LLMs often exhibit limitations in complex reasoning and multi-step planning when relying solely on internal parametric knowledge. To bridge this gap, recent research has increasingly integrated external tool use and explicit planning strategies to augment model capabilities~\cite{wang2024survey}. Unlike passive inference, agentic systems typically maintain intermediate states, interleave reasoning with actions, and invoke external tools to acquire information or execute computations, enabling more modular decision workflows. Prominent methodologies typically follow structured paradigms, such as the iterative data-flow approach or the plan-and-execute paradigm, where the model generates a comprehensive plan before invoking external tools. While effective for standard tasks, these methods often rely on rigid, predefined pipelines or static templates. This dependence significantly limits their flexibility, making them brittle when facing open-ended problems that require adaptive strategy adjustment.

To overcome the constraints of static workflows, the field is advancing towards dynamic agent frameworks capable of autonomous planning and on-the-fly tool selection. Representative systems integrate tool use, including ReAct~\cite{yao2023react}, Toolformer~\cite{schick2023toolformer}, and WebGPT~\cite{nakano2021webgpt}, which improve reliability by grounding outputs in tool-mediated observations. Moreover, modern agent frameworks extend this paradigm by introducing advanced modules like planning~\cite{wang2023describe}, memory~\cite{packer2023memgpt}, and iterative tool coordination~\cite{shinn2023reflexion}, supporting robust multi-step execution~\cite{erdogan2025plan}. However, directly adapting these text-centric paradigms to data-intensive tasks remains challenging, as they often lack the rigor required for precise numerical manipulation and logical verification. Building on these autonomous concepts, our work extends the paradigm to fully programmatic execution~\cite{cheng2025agentr1trainingpowerfulllm}. By internalizing the executable environment as an external cognitive organ, we shift the reasoning paradigm from imprecise textual simulation to exact code-based computation. Unlike general tool-use agents, we focus on training a smaller-scale model to autonomously write and execute code within a secure sandbox. This approach addresses the largely unexplored challenge of enabling stable, code-driven reasoning and execution in resource-constrained environments, while  establishing a foundation for uncertainty-aware interaction.

\subsection{Uncertainty Quantification}
Effective uncertainty quantification is a prerequisite for reliable agentic systems, serving as a critical indicator for distinguishing between aleatoric uncertainty  and epistemic uncertainty~\cite{yadkori2024believe,kapoor2024large}. While early methods relied on token-level log-likelihoods, recent research highlights that these often fail to capture the true reliability of outputs due to semantic equivalence, where different phrasings represent the same meaning~\cite{grewal2024improving,ling2024uncertainty}. To address this, semantic entropy was proposed to measure uncertainty at the meaning level rather than the lexical level, offering a robust tool for detecting hallucinations~\cite{farquhar2024detecting}. Further advancements include kernel language entropy (KLE)~\cite{nikitin2024kernel}, which utilizes positive semidefinite kernels to encode semantic similarities for fine-grained estimation, and semantic embeddings approaches that model semantics as latent variables to reduce sensitivity to irrelevant words~\cite{kuhnsemantic}. Additionally, models can be trained to verbalize their confidence explicitly; for instance, SaySelf~\cite{xu2024sayself,xiongcan} teaches LLMs to express fine-grained confidence estimates alongside self-reflective rationales~\cite{ yao2022react, shinn2023reflexion}.

Beyond static evaluation, UQ plays a pivotal role in monitoring and controlling the intermediate steps of complex reasoning chains~\cite{ye2024benchmarking,shorinwa2025survey}. Unlike prompt-level assessment, CoT-UQ~\cite{zhang2025cot} introduces a response-wise framework that integrates UQ into the chain-of-thought (CoT)~\cite{wei2022chain} process, assessing the importance of keywords in each reasoning step to detect potential errors early. This step-wise monitoring is crucial for identifying unreliable generation trajectories, enabling mechanisms to prune high-uncertainty plans before execution, similar to the exploration strategies in tree of thoughts (ToT)~\cite{yao2023tree}. Furthermore, fine-grained token-level methods like claim conditioned probability (CCP)~\cite{fadeeva2024fact} remove the impact of surface form uncertainty to specifically measure the uncertainty of atomic claims, serving as an effective tool for fact-checking. Other approaches, such as APRICOT~\cite{ulmer2024calibrating}, train an auxiliary model to predict the main model's confidence based on input and output alone, facilitating calibration without interfering with generation.
To synthesize reliable final answers from multiple reasoning paths, UQ serves as a weighting mechanism for aggregation. The standard self-consistency paradigm~\cite{wangself} improves performance by sampling diverse reasoning paths and applying majority voting, leveraging the intuition that complex problems admit multiple ways of thinking leading to a unique correct answer. Advanced approaches extend this: Universal self-consistency (USC)~\cite{chen2024universal} leverages the LLM itself to select the most consistent answer among candidates, making it applicable to free-form generation tasks~\cite{linteaching}. Furthermore, Conformal Language Modeling applies conformal prediction to calibrate stopping and rejection rules, constructing prediction sets that guarantee coverage of acceptable responses~\cite{quachconformal}. 

%% file: 3-Preliminaries.tex
\section{Preliminaries}
In this section, we first define the table reasoning problem and formalize its key components. We then present the core ideas and underlying principles of TableMind++ that guide our approach.
\subsection{Task Formulation}
We consider the task of table reasoning as a structured decision-making problem grounded in tool-augmented language modeling. Each instance is represented as a tuple $(T, Q, A)$, where $T$ denotes a structured table, $Q$ is a natural language query (e.g., a question or a factual statement), and $A$ is the corresponding ground-truth answer. Unlike conventional QA settings, table reasoning often involves multi-step computation, intermediate result verification, and error correction. Thus, the task requires not only semantic understanding of $T$ and $Q$, but also precise numerical reasoning, which necessitates intermediate state tracking, external tool usage, and dynamic adjustment of the reasoning path.

Formally, the goal is to learn a robust policy $\pi$ that, given the initial inputs $(T, Q)$, generates a sequence of executable intermediate reasoning steps $\{s_k\}_{k=1}^{K}$. Crucially, distinct from standard approaches that solely aim for a deterministic output, our framework targets reliable uncertainty-aware reasoning. The system produces a final output tuple $(\hat{A}, u) = \mathcal{R}(\{s_k\}_{k=1}^{K})$, where $\hat{A}$ is the predicted answer and $\textbf{\textit{u}}$ represents the quantified uncertainty (or confidence) of the reasoning trajectory. Here, $\mathcal{R}$ denotes an aggregation function (e.g., weighted voting) that synthesizes the final result from generated paths. The objective is to jointly maximize the probability of predicting the correct $A$ while ensuring that $u$ is well-calibrated with the true correctness likelihood.
\subsection{Design Principles}
LLMs demonstrate strong capabilities in understanding natural language, making them valuable tools for the complex task of table reasoning. However, current research presents a distinct dilemma, forcing a choice between two insufficient paradigms. Direct reasoning methods often lack mechanisms for precise computation and external verification, leading to frequent calculation errors. Conversely, rigid tool-augmented pipelines are often heavily reliant on massive, proprietary models and predefined templates, lacking the flexibility for dynamic reflection. To bridge these gaps and construct a system that is not only capable but also reliable, we design TableMind++ based on three core principles.

First, we aim to train an autonomous programmatic agent rather than merely constructing a static workflow. Instead of relying on external prompts, the agent must possess the ability to dynamically decompose problems, write executable code in a sandbox, and critically reflect on execution feedback to self-correct errors. Second, we emphasize strategic efficiency via policy optimization. We adhere to the principle that high-performance reasoning should not depend exclusively on scaling up model size. Small, open-source models can achieve expert-level capability if their reasoning policies are explicitly optimized, driving our use of reinforcement learning to align the model's strategy with successful outcomes.
Third, and most critically for this extended framework, we enforce Trustworthiness via Uncertainty Awareness. We posit that autonomy without guardrails introduces risk, as models may confidently pursue erroneous paths. Therefore, a robust agent must actively monitor its uncertainty during the planning phase to prune hallucinated strategies and calibrate its final answer during the inference phase.

%% file: 4-SystemDesign.tex
\begin{figure*}[t]
    \centering
    \includegraphics[width=1\linewidth]{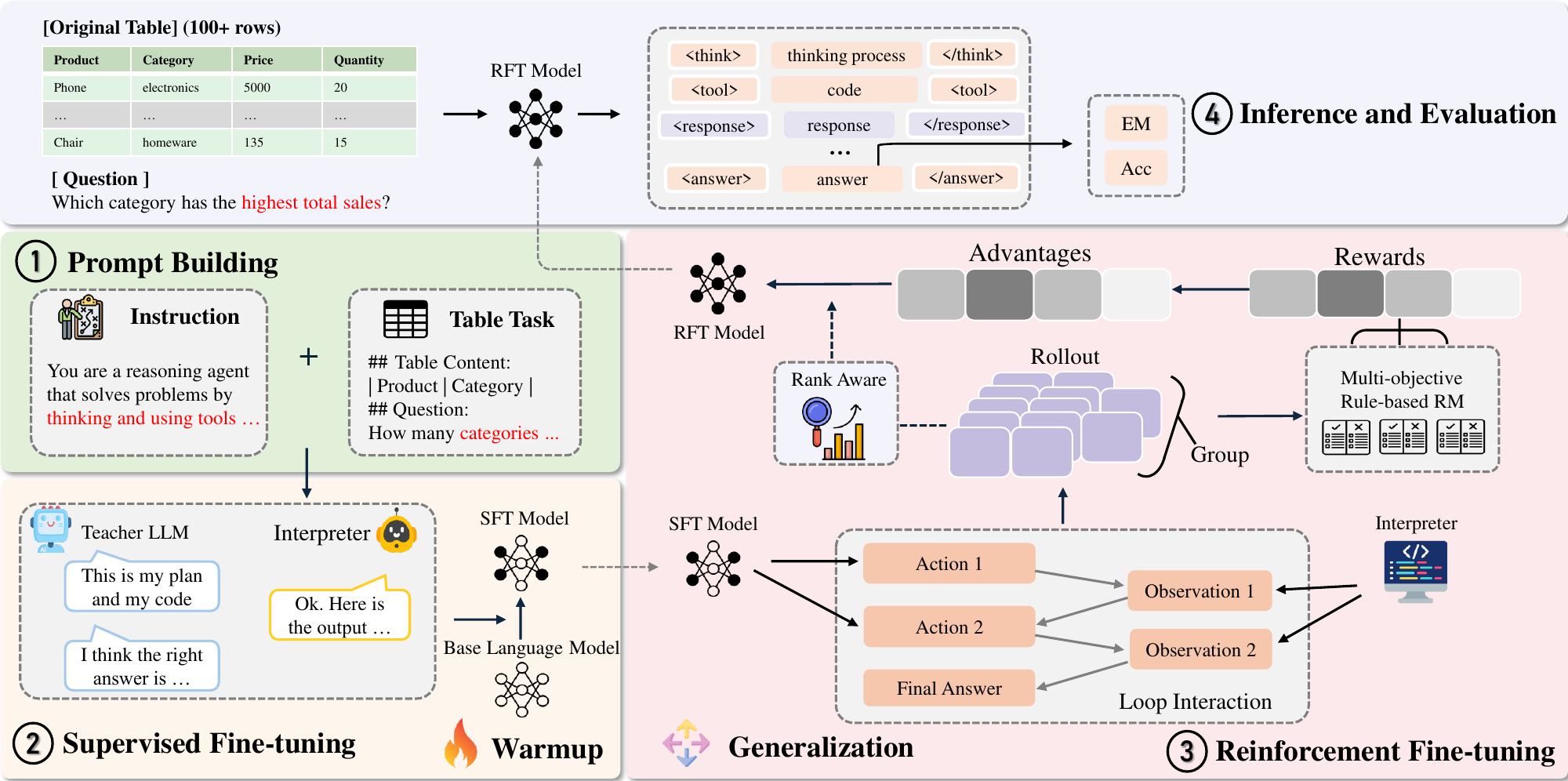}
    \caption{The overall training pipeline for TableMind. The process begins with Prompt Building and SFT to warm up the model by providing it with a strong initial policy. Subsequently, RFT with RAPO is applied to significantly enhance its generalization capability. The final model is then deployed for Inference and evaluated for accuracy.}
    \label{fig:frameDiagram}
\end{figure*}
\section{TableMind}
This section presents the training pipeline of TableMind. It begins with the prompt template design, followed by the two-stage training strategy, and concludes with inference and evaluation.
\subsection{Overview}
As illustrated in Figure~\ref{fig:frameDiagram}, the TableMind framework is built upon a standardized Prompt Template that unifies diverse table inputs and task descriptions, guiding the model to interleave natural language reasoning with tool usage. The framework is established through a principled two-stage training strategy. First, the Supervised Fine-Tuning (SFT) stage bootstraps the model with fundamental tool-invocation and reasoning capabilities using high-quality trajectories distilled from a teacher model. Subsequently, the Reinforcement Learning (RL) stage enables the agent to autonomously explore and refine its strategies. To guarantee optimization stability, we introduce the Rank-Aware Policy Optimization (RAPO) algorithm coupled with a multi-perspective reward function, which effectively aligns model confidence with trajectory quality. During inference, the agent executes iterative, multi-turn reasoning to derive precise answers.

\subsection{Prompt Template Design}
We propose a structured prompt template to ensure consistent and reliable reasoning for table-based tasks, aiming to mitigate ambiguity in table comprehension and provide a unified interface for reasoning and tool usage. The prompt integrates high-level reasoning guidance with task-specific context to address the inherent complexity of table reasoning, which demands coordinated data interpretation, problem decomposition, and tool invocation. Concretely, it comprises a general instruction block that defines the agent's role and enforces a rigorous thought-action-observation workflow, and a table task block that presents the serialized table schema and the corresponding query. This design effectively mitigates format hallucinations, yielding more stable reasoning trajectories and precise tool usage across diverse tasks.

\subsection{Supervised Fine-tuning} The SFT stage primarily serves to initialize the agent’s interaction behavior by distilling high-quality trajectories and aligning the model with a structured reasoning paradigm.

\subsubsection{Data Collection} We construct a high-quality dataset of reasoning trajectories for SFT using an iterative, multi-turn synthesis framework. Leveraging a knowledge distillation~\cite{hinton2015distilling} approach, we transfer the programmatic reasoning capabilities of an expert model into structured training samples. Specifically, the expert model is prompted to generate a plan and an initial executable action, which is then executed in a local sandbox. The resulting execution feedback is returned to the model as an observation, enabling it to reflect on the state, refine its plan, and generate subsequent actions. This interactive loop continues until the task is completed and a final answer is derived. To ensure data quality, all collected trajectories undergo a rigorous validation process: we compare the final output against a predefined ground-truth answer and retain only those trajectories that successfully reach the correct solution, thereby filtering out hallucinations or execution failures.

\subsubsection{Warm-up Training Stage} The SFT stage functions as a warm-up phase that standardizes the model’s output format, enables it to internalize the syntax of the planning-action-reflection loop, and establishes a foundational policy for generating syntactically valid and logically coherent code. This initial phase acts as a crucial preparatory step, substantially lowering the optimization overhead for the subsequent reinforcement learning stage. However, while SFT successfully imparts foundational instruction-following capabilities, the resulting reasoning behavior often lacks robustness. It tends to reflect surface-level pattern memorization from the training data and exhibits limited generalization to out-of-distribution tasks. Consequently, the reasoning ability induced by SFT can be viewed as a form of rote memory, necessitating further refinement through reinforcement learning to achieve adaptive generalization.
\subsection{Reinforcement Fine-tuning} 
Building upon the foundational capabilities established during SFT, the RFT stage focuses on improving generalization by optimizing the agent via carefully designed reward functions and a rigorous policy optimization objective.

\subsubsection{Multi-perspective Reward Design} We construct a multi-perspective reward mechanism comprising three distinct components. This design effectively guides the agent's exploration of tool-use strategies and enhances the sample efficiency of RLVR~\cite{yue2025does}.
To guarantee the structural integrity of the generated reasoning paths, we define a format reward $R_{\mathrm{format}}$. This reward enforces strict adherence to the thought-action-observation protocol, ensuring the model produces syntactically valid reasoning trajectories.
\begin{equation}
R_{\mathrm{format}} =
\begin{cases}
1, & \text{if output format is valid}, \\
0, & \text{otherwise}.
\end{cases}
\label{eq:format_reward}
\end{equation}

A binary reward is assigned based on whether the output strictly conforms to the required structural constraints.
Serving as the definitive metric of utility, the accuracy reward $R_{\mathrm{acc}}$ directly evaluates the correctness of the agent's final derived solution. The reward is defined as:

\begin{equation}
R_{\mathrm{acc}} =
\begin{cases}
1, & \text{answer matches the ground truth}, \\
0, & \text{otherwise}.
\end{cases}
\label{eq:accuracy_reward}
\end{equation}

Regarding the accuracy reward $R_{\mathrm{acc}}$, the metric varies by task type: we employ exact match for Table QA and label accuracy for Fact Verification. Furthermore, to incentivize optimal tool utilization, we introduce an auxiliary reward component, $R_{\mathrm{tool}}$, designed with an implicit curriculum mechanism that promotes early-stage exploration while enforcing execution efficiency:
\begin{equation}
R_{\mathrm{tool}} = e^{-\rho s} \left( \beta \cdot I_{\mathrm{success}} 
- C \cdot (N_{\mathrm{turns}})^{2} \right),
\label{eq:tool_reward}
\end{equation}
where $s$ is the number of global training steps, $\mathbb{I}(\cdot)$ is the indicator~function that returns $1$ if at least one tool call in the trajectory is successful, $\beta$ is the positive base reward, $N_{\mathrm{turns}}$ is the number of tool turns, and $C$ is a penalty coefficient. The total reward for each reasoning trajectory is the sum of the above components.

\subsubsection{RAPO}
We introduce Rank-Aware Policy Optimization (RAPO), a refined group-based policy gradient algorithm designed to enhance learning stability. RAPO identifies misaligned trajectories and boosts their learning signals through rank-aware advantage weighting. Building upon the GRPO~\cite{shao2024deepseekmath} framework, RAPO incorporates three strategic enhancements to address the limitations of existing methods. First, following recent findings~\cite{hu2025open, liu2025understanding}, we eliminate the KL divergence penalty, thereby liberating the model from the constraints of the reference policy and expanding the exploration space. Second, we employ a token-level policy gradient loss to normalize updates,  mitigating the length bias where longer answers excessively influence the gradient. Finally, we adopt a modified clipping strategy, which promotes generation diversity and prevents entropy collapse. The objective function is formalized as:
\begin{equation}
\begin{aligned}
J_{\mathrm{RAPO}}(\theta) 
&= \mathbb{E}_{q, \{o_i\}_{i=1}^G \sim \pi_{\theta_{\mathrm{old}}}} \bigg[\quad \frac{1}{\sum_{i=1}^G |o_i|} \sum_{i=1}^G 
\sum_{t=1}^{|o_i|} 
\min\big( r_{i,t}(\theta) A'_i,\; \\
&\mathrm{clip}(r_{i,t}(\theta), 1-\epsilon_{\mathrm{low}}, 1+\epsilon_{\mathrm{high}}) A'_i \big)
\bigg],
\end{aligned}
\end{equation}
where $r_{i,t}(\theta)$ represents the likelihood ratio between the current and old policies:
\begin{equation}
\begin{aligned}
r_{i,t}(\theta) = \frac{\pi_{\theta}(o_{i,t} \mid q, o_{i,<t})}{\pi_{\theta_{\mathrm{old}}}(o_{i,t} \mid q, o_{i,<t})}.
\end{aligned}
\end{equation}

Building upon this framework, we substitute the standard group-normalized advantage with a dynamically weighted, rank-aware advantage $A'_i$. This modification ensures that the gradient signal for each token scales not only with the trajectory’s quality but also with the severity of the misalignment between the model’s confidence and that quality:
\begin{equation}
\begin{aligned}
A'_i = \gamma_i \cdot \frac{R_i - \mathrm{mean}(\{R_j\}_{j=1}^G)}{\mathrm{std}(\{R_j\}_{j=1}^G)}.
\end{aligned}
\end{equation}

The core mechanism of RAPO lies in identifying and leveraging learning signals from misaligned trajectory pairs. A misalignment is defined as an instance where the model assigns higher confidence to a lower-reward trajectory ($o_l$) compared to a higher-reward trajectory ($o_w$) within the same group. Here, confidence is quantified by the length-normalized log-probability of the sequence, denoted as $\log P(o_i)$.To capture these misalignments, we introduce a pairwise weighting coefficient $\gamma_{w,l}$. This term specifically targets winner-loser pairs where the model's confidence ranking contradicts the reward ranking. When the winner's log-probability fails to exceed that of the loser, the pair is deemed misaligned, and $\gamma_{w,l}$ is increased to enforce a stronger optimization signal. The factor is defined as:
\begin{equation}
\gamma_{w,l} = 1 + \alpha \cdot \mathbb{I}[\log P(o_w) < \log P(o_l)],
\end{equation}
where $\mathbb{I}[\cdot]$ is the indicator function and $\alpha$ is a hyperparameter controlling the intensity of the re-weighting.

\subsection{Towards Uncertainty-Aware Reasoning}
Table reasoning is critical in high-stakes domains such as quantitative finance, medical data analysis, and scientific research. In these contexts, reliability is paramount, as even minor errors can precipitate severe consequences. However, LLMs inherently operate probabilistically, making them susceptible to hallucinations. This challenge is further compounded in autonomous agentic frameworks like TableMind. Given the multi-turn nature of the reasoning process, uncertainty tends to propagate and accumulate across steps. A minor deviation in the initial planning phase can cascade into catastrophic downstream errors.

While ensemble methods such as self-consistency (SC)~\cite{wangself} can effectively suppress stochastic noise by aggregating multiple sampled paths, applying naive SC to multi-turn agentic workflows is highly inefficient. Generating numerous full-length interaction trajectories, each requiring multiple external tool invocations, incurs substantial latency and resource overhead. Although our proposed RAPO training strategy aligns model confidence with trajectory quality, explicitly integrating uncertainty quantification (UQ) during the inference phase emerges as an imperative extension. By quantifying uncertainty at each step, the agent can dynamically prune unreliable branches and calibrate final answers, rather than relying on indiscriminate sampling. This motivates the evolution of our framework to TableMind++, which incorporates UQ-enhanced inference mechanisms to guarantee robust decision.

%% file: 4.5-SystemDesign.tex
\section{TableMind++} \label{sec:tablemind_pp} This section presents TableMind++, an advanced inference framework engineered for high-stakes reliability. By rigorously quantifying both epistemic and aleatoric uncertainties, we transition the agent from static generation to dynamic, risk-aware reasoning.

\subsection{Overview of TableMind++} 
\begin{figure*}[t]
    \centering
    \includegraphics[width=1\linewidth]{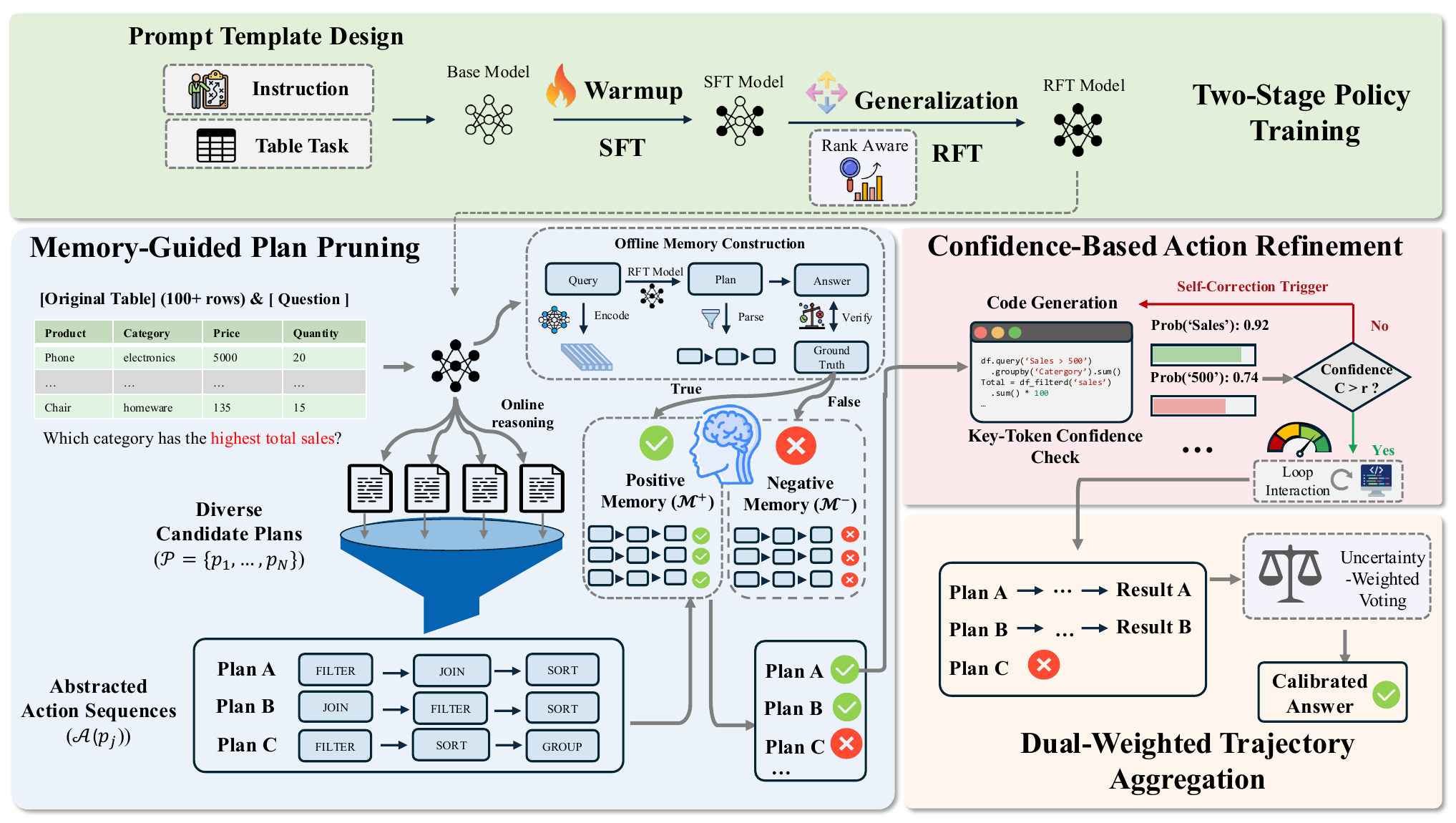}
\caption{The architecture of TableMind++. The framework integrates two-stage policy optimization with an uncertainty-aware inference pipeline. During inference, the system sequentially employs memory-guided plan pruning to filter logical errors, confidence-based action refinement to correct syntactic noise, and dual-weighted trajectory aggregation to derive a calibrated final consensus.}
    \label{fig:frameDiagram++}
\end{figure*}

While TableMind establishes solid reasoning capabilities through SFT and RFT, deploying autonomous agents in critical domains demands strict reliability guarantees. A fundamental challenge lies in the probabilistic nature of LLMs, which introduces significant uncertainty into the reasoning process. To address this rigoroulsy, we decouple the uncertainty in agentic reasoning into two distinct categories: epistemic uncertainty and aleatoric uncertainty.
Epistemic uncertainty arises from the model's structural ambiguity regarding the optimal logical path for a specific query, rather than a mere lack of factual knowledge. In the planning phase, high epistemic uncertainty manifests as logical hallucinations, where the model generates plans that are linguistically plausible but structurally unsound. Crucially, this issue cannot be resolved by simple sampling strategies; instead, it necessitates structural validation against historical reasoning trajectories to prune deviating paths that fail to align with proven logical patterns.
Conversely, aleatoric uncertainty stems from the inherent stochasticity in the token generation process, representing unavoidable data or generation noise. Even when operating with a correct high-level plan, this noise can precipitate syntactic errors, incorrect API calls, or hallucinated arguments during the code execution phase, which are directly mirrored by the instability of the token-level probability distribution.
As illustrated in Figure \ref{fig:frameDiagram++}, we propose TableMind++, a dynamic inference framework employing a hybrid quantification strategy. We specifically design Memory-Guided Plan Pruning to mitigate epistemic uncertainty by anchoring reasoning to proven historical strategies, and Confidence-Based Action Refinement to manage aleatoric uncertainty by monitoring real-time token stability.

\subsection{Memory-Guided Plan Pruning}
\label{sec:plan_pruning}

In the planning phase, the output space comprises open-ended natural language. The primary risk here is epistemic uncertainty which arises from the insufficiency of the model's internal knowledge to generalize across diverse table schemas and question types. Consequently, the model may confidently hallucinate plausible but logically flawed plans. Since the model's internal confidence is often miscalibrated in such scenarios, we rely on an external epistemic anchor, a dual-memory bank, to validate the reasoning logic.
We propose a memory-guided plan pruning strategy. Specifically, for each input query, we first sample a set of diverse candidate plans to cover the potential reasoning space. Our method then operates on abstracted Action Sequences to filter these candidates based on their sequential proximity to historical successes and known failure patterns.

\subsubsection{Dual-Memory Construction }

We construct an offline memory bank $\mathcal{M}$ by aggregating reasoning trajectories generated by the optimized TableMind agent itself. Specifically, we execute the converged model across the entire training dataset to collect self-generated reasoning paths. This self-referential approach ensures that the memory effectively captures the model's inherent behavioral distributions, encompassing both robust reasoning strategies and characteristic error modes.
To facilitate generalization, we employ a semantic parser to canonicalize free-form planning text into a sequence of logical primitives from a predefined set $\mathbb{O} = \{\texttt{FILTER}, \texttt{GROUP}, \texttt{AGGREGATE}, \dots\}$. The parser identifies reasoning keywords, for instance, mapping ``remove rows'' or ``keep only'' to \texttt{FILTER}, while strictly discarding schema-specific arguments such as column names and numerical literals. This abstraction ensures that the resulting sequence represents the pure underlying logical structure of the solution, independent of specific data values.
Formally, each memory entry is stored as a comprehensive tuple $(x_j, p_j, E(x_j), \mathcal{A}(p_j))$. Here, $x_j$ and $p_j$ denote the raw natural language query and plan, preserved for contextual reference. $E(x_j)$ represents the dense vector embedding generated by a pre-trained sentence encoder $E$, serving as the retrieval key. $\mathcal{A}(p_j)$ is the abstracted action sequence that serves as the structural template for pruning. We categorize these entries into two distinct subsets: Positive Memory ($\mathcal{M}^+$) and Negative Memory ($\mathcal{M}^-$). The former stores trajectories that successfully derived the correct answer, while the latter retains deceptive trajectories, defined as paths that were executable but yielded incorrect results (e.g., substituting summation for averaging), to serve as explicit negative constraints.

\subsubsection{Memory-Guided Pruning}
During inference, we employ a retrieve-then-filter mechanism to validate the generated reasoning paths. First, for a new query $x$, we explore the potential reasoning space by generating a candidate set $\mathcal{P} = \{p_1, \dots, p_N\}$ via temperature sampling. Simultaneously, we encode the query into a dense vector $E(x)$ using the same encoder and perform a nearest neighbor search across the constructed memory bank. We select the top-$K$ historical instances based on the cosine similarity between the current query embedding $E(x)$ and the stored embeddings $E(x_j)$. The action sequences associated with these retrieved instances are then partitioned into a set of positive prototypes $\mathbb{S}_{pos}$ and negative prototypes $\mathbb{S}_{neg}$.
To evaluate the structural validity of the sampled candidates, we convert each plan $p_i \in \mathcal{P}$ into its action sequence $\mathcal{A}(p_i)$. We then calculate its sequential distance to the nearest positive prototype ($D^+$) and the nearest negative prototype ($D^-$) using the Levenshtein Edit Distance~\cite{levenshtein1965binary}:
\begin{equation}
\begin{aligned}
D^+(p_i) &= \min_{\mathcal{A}_{ref} \in \mathbb{S}_{pos}} \operatorname{Lev}(\mathcal{A}(p_i), \mathcal{A}_{ref}) \\
D^-(p_i) &= \min_{\mathcal{A}_{ref} \in \mathbb{S}_{neg}} \operatorname{Lev}(\mathcal{A}(p_i), \mathcal{A}_{ref})
\end{aligned},
\end{equation}
where $\operatorname{Lev}(\cdot, \cdot)$ denotes the edit distance function, which quantifies the minimal number of insertions, deletions, or substitutions required to transform the candidate sequence into the reference pattern.
Finally, we calculate a contrastive score $S_{con}(p_i)$ for each plan to quantify its structural preference towards historical success over failure:
\begin{equation}
S_{con}(p_i) = D^-(p_i) - D^+(p_i),
\end{equation}
where a higher score indicates that the plan is significantly closer to verified strategies while maintaining a safe distance from known pitfalls. Based on this score, we rank all candidates in descending order and retain only the top portion determined by a ratio $\rho$ to form the refined set $\mathcal{P}^*$. This mechanism ensures robustness by prioritizing the relatively best logical structures, effectively filtering out hallucinations that resemble known failure patterns while preserving the diversity of potential solutions.

\subsection{Confidence-Based Action Refinement}
\label{sec:action_refinement}

Upon establishing a structurally valid plan, the agent transitions to the execution phase to generate executable code. At this stage, the primary challenge shifts to aleatoric uncertainty, which stems from the inherent stochasticity of the generative process. Even when the high-level intent is correct, stochastic sampling can introduce syntactic inconsistencies in critical details, such as variable names or numerical literals.
A major obstacle in quantifying this uncertainty is probability dilution. This phenomenon arises because the boilerplate syntax of code—such as keywords and assignment operators—is highly deterministic and typically generated with near-perfect probability. Consequently, these high-confidence tokens artificially inflate the average sequence score, potentially masking low-confidence hallucinations embedded within critical logical segments.

To quantify this uncertainty without the bias of probability dilution, we compute the generation confidence exclusively over semantically significant tokens. Let $a$ denote the generated code sequence comprising a series of tokens $a_1$ through $a_L$. We define a subset of indices $\mathcal{K}$ within the range $\{1, \dots, L\}$ that corresponds to high-information components identified via lexical analysis. These components specifically include variable identifiers, function names, and numerical or string literals. By excluding deterministic syntactic markers, the refined confidence score $\mathcal{C}(a)$ is calculated solely over these critical regions:
\begin{equation}
\mathcal{C}(a) = \exp \left( \frac{1}{|\mathcal{K}|} \sum_{i \in \mathcal{K}} \log P_\theta(a_i \mid x, a_{<i}) \right),
\end{equation}
where $P_\theta$ represents the probability distribution of the model parameterized by weights $\theta$, and $a_{<i}$ denotes the sequence of tokens generated prior to step $i$.
We interpret $\mathcal{C}(a)$ as a direct measure of aleatoric confidence. Leveraging this signal, we employ a conditional execution mechanism. Instead of immediately executing the generated code, if $\mathcal{C}(a)$ falls below a threshold $\tau_{\text{code}}$, the system halts the execution flow and triggers a refinement cycle. Specifically, the model is prompted to self-correct the generated artifacts by reviewing and regenerating the code, with a focus on resolving ambiguities in the identified low-probability identifiers or literals. This pre-execution intervention prevents noisy code from polluting the context window with error tracebacks.

\subsection{Dual-Weighted Trajectory Aggregation}
Following the pruning and refinement phases, we obtain a set of verified trajectories $\mathcal{T}^* = \{(p_i, \mathbf{h}_i, y_i)\}_{i=1}^{M}$. Here, $M$ denotes the total number of retained candidates, and $\mathbf{h}_i$ represents the complete execution history, encompassing potentially multiple rounds of action generation and environmental observations such as iterative debugging or multi-step reasoning steps. Correspondingly, $y_i$ denotes the final derived answer. While the previous steps effectively mitigate individual epistemic and aleatoric uncertainties, relying on a single trajectory remains susceptible to stochastic variance. To derive a proficient final answer, we adopt a weighted voting scheme that synthesizes the consensus across high-quality reasoning paths.
Unlike standard majority voting which treats all trajectories equally, we assign a confidence weight $w_i$ to each trajectory by synthesizing its structural validity and the cumulative certainty of its execution steps. Specifically, we combine the normalized contrastive score from the planning phase and the refined confidence score averaged over the generated actions within the history $\mathbf{h}_i$:
\begin{equation}
w_i = \sigma(S_{con}(p_i)) \cdot \mathcal{C}(\mathbf{h}_i),
\end{equation}
where $\sigma(\cdot)$ denotes the sigmoid function which normalizes the unbounded contrastive score into the probability interval between 0 and 1. This composite weight effectively prioritizes trajectories that demonstrate logical soundness through structural similarity to verified prototypes, while simultaneously maintaining high syntactic confidence throughout the interaction loop.
Finally, we group the trajectories by their execution results and select the answer $\hat{y}$ that maximizes the accumulated weight:
\begin{equation}
\hat{y} = \operatorname*{arg\,max}_{y \in \mathcal{Y}} \sum_{i=1}^{M} \mathbb{I}(y_i = y) \cdot w_i,
\end{equation}
where $\mathcal{Y}$ is the set of unique answers produced by the trajectories, and $\mathbb{I}(\cdot)$ is the indicator function. This aggregation mechanism ensures that the final output represents the most reliable consensus, reinforced by both memory-retrieved logic and execution-time certainty.

%% file: 5-Experiments.tex
\definecolor{color1}{HTML}{DDEBF7} 
\definecolor{color2}{HTML}{BDD7EE} 
\definecolor{color3}{HTML}{5B9BD5} 
\definecolor{color4}{HTML}{2F75B6} 
\definecolor{color5}{HTML}{204E79} 

\newcommand{\legenditem}[2]{%
    \colorbox{#1}{\rule{0pt}{2pt}\rule{3pt}{0pt}}\hspace{0.5em}#2%
}

\input{table/4-dataset}

\section{Experiments}
In this section, we validate TableMind++ through comprehensive experiments. We first outline the experimental setup and main results, followed by ablation studies on training and inference mechanisms. Finally, we present in-depth analyses of model dynamics, efficiency, and sensitivity, concluded by qualitative case studies.
\subsection{Experimental Setup}
\subsubsection{Datasets.} 
Table~\ref{tab:dataset_stats} presents detailed dataset statistics. We evaluate TableMind++ on comprehensive table reasoning benchmarks, categorized into in-domain and out-of-domain settings to fully assess the model's performance and generalization capabilities. 
For training, we construct a mixed dataset by randomly sampling 3,500 instances each from the WikiTQ and TabFact training sets, and 1,000 from TabMWP. 
For in-domain evaluation, we utilize three widely used public datasets: WikiTQ~\cite{pasupat2015compositional}, TabMWP~\cite{lu2022dynamic}, and TabFact~\cite{chen2019tabfact}. 
WikiTQ focuses on open-domain question answering where models generate concise answers from Wikipedia tables and textual context. 
TabMWP targets mathematical reasoning, involving multi-step numerical calculations and quantitative relationship reasoning. 
TabFact addresses fact verification by determining whether a textual statement is supported by a given table. 
To further evaluate the robustness and transferability of TableMind++, we extend our experiments to two out-of-domain datasets: HiTab~\cite{cheng2021hitab} and FinQA~\cite{chen2021finqa}. 
HiTab introduces hierarchical tables requiring the understanding of complex header dependencies, while FinQA targets expert-level financial reasoning with multi-step calculations over both tables and text. 
All evaluations are conducted on the official test sets of the respective datasets to ensure consistency and fair comparison with prior work.

\subsubsection{Baselines.} 
We compare TableMind++ against a comprehensive set of strong baselines, encompassing training-free frameworks, tuning-based models, and state-of-the-art general LLMs. 
For training-free methods, we employ \textbf{Tab-CoT}~\cite{jin2023tab}, which prompts LLMs to model complex reasoning processes step-by-step; \textbf{PoTable}~\cite{mao2024potable}, which utilizes a plan-then-execute framework with external tools; and \textbf{Chain-of-Table}~\cite{wang2024chain}, which introduces an evolving table paradigm where the table serves as input for subsequent operations. 
Regarding tuning-based approaches, we select \textbf{TableLlama}~\cite{zhang2023tablellama} and \textbf{TableGPT2}~\cite{su2024tablegpt2} for their specialized instruction-tuning and semantic encoding capabilities, as well as \textbf{Table-R1}~\cite{yang2025table}, which leverages reinforcement learning with verifiable rewards. 
To fully assess current capabilities, we also evaluate powerful general LLMs via direct prompting, including proprietary models such as GPT-4.1, GPT-5~\cite{gpt4}, and the Gemini-2.5~\cite{gemini} series, alongside competitive open-source models like Deepseek-R1~\cite{deepseek} and Qwen2.5-72B-Instruct~\cite{qwen2}. 
Finally, we explicitly include Qwen3-8B~\cite{qwen3} as the backbone baseline to isolate the specific contributions of our training framework, and our previous work \textbf{TableMind}~\cite{jiang2025tablemind} to demonstrate the incremental performance advancements achieved by TableMind++.
\subsubsection{Implementation Details.}

We utilize Qwen3-8B as the backbone model. The training pipeline consists of two stages:
During SFT, we warm up the model on 200 synthetic samples for 1 epoch with a learning rate of $1 \times 10^{-6}$.
Subsequently, in the RFT stage, we implement the RAPO algorithm using the \texttt{verl} framework, employing \texttt{vLLM} for efficient generation. The reward hyperparameters in Eq.~\eqref{eq:tool_reward} are configured as $\rho = 0.05$, $C = 0.01$, $\beta = 0.5$, and the group size is set to $G = 8$. Training uses a batch size of 128 and a learning rate of $1 \times 10^{-6}$. The policy temperature is set to 1.0, with a maximum of 3 tool-calling turns and a response limit of 2048 tokens per turn. All training experiments are conducted on a cluster of 4 NVIDIA A800 GPUs.
TableMind++ operates as an inference-time wrapper around the optimized model. For memory-guided pruning, we employ bge-m3~\cite{chen2024bge}  as the dense encoder. For each query, we sample $N=16$ candidate reasoning paths (temperature $= 1.0$) and retrieve the top-$K=5$ historical prototypes from the constructed memory bank. The pruning retention ratio is set to $\rho = 0.5$, preserving only the top-ranked candidates for execution.For Confidence-Based Action Refinement, we set the token stability threshold to $0.9$ to trigger self-correction. To ensure a strictly fair comparison, we standardize the final output to a single pass (Pass@1) across all baselines.

\input{table/1-main-result}

\subsection{Main Results}
To evaluate the efficacy of our proposed framework, we conduct a comprehensive comparison against competitive baselines across diverse benchmarks, spanning both in-domain and out-of-domain tasks. The results summarized in Table~\ref{tab:main result} demonstrate that TableMind++ establishes a new state-of-the-art, consistently outperforming strong proprietary LLMs such as GPT-5 and Gemini-2.5-flash, as well as leading open-source models.
Notably, on complex out-of-domain datasets like HiTab and FinQA, TableMind++ exhibits remarkably robust generalization capabilities, significantly surpassing specialized baselines like TableLlama and Table-R1. Unlike training-free methods such as Tab-CoT and Chain-of-Table, which rely heavily on prompt engineering, our tuning-based approach successfully internalizes reasoning capabilities. Moreover, while Table-R1 utilizes rejection sampling, it still falls short of our framework; TableMind outperforms it by substantial margins on tasks such as TabFact, validating the effectiveness of our RLVR training paradigm.
Furthermore, even compared to powerful general reasoning models like Deepseek-R1, TableMind++ maintains a clear advantage, particularly in specialized financial reasoning contexts. The analysis also highlights the progression from TableMind to TableMind++; while TableMind already secures a leading position in most categories, TableMind++ further pushes the performance boundaries. This improvement is especially evident in logical reasoning and hierarchical table understanding, confirming that our advanced inference strategies effectively minimize calculation errors and hallucinations in complex structural reasoning.

\begin{figure*}[t]
    \centering
    \includegraphics[width=1\linewidth]{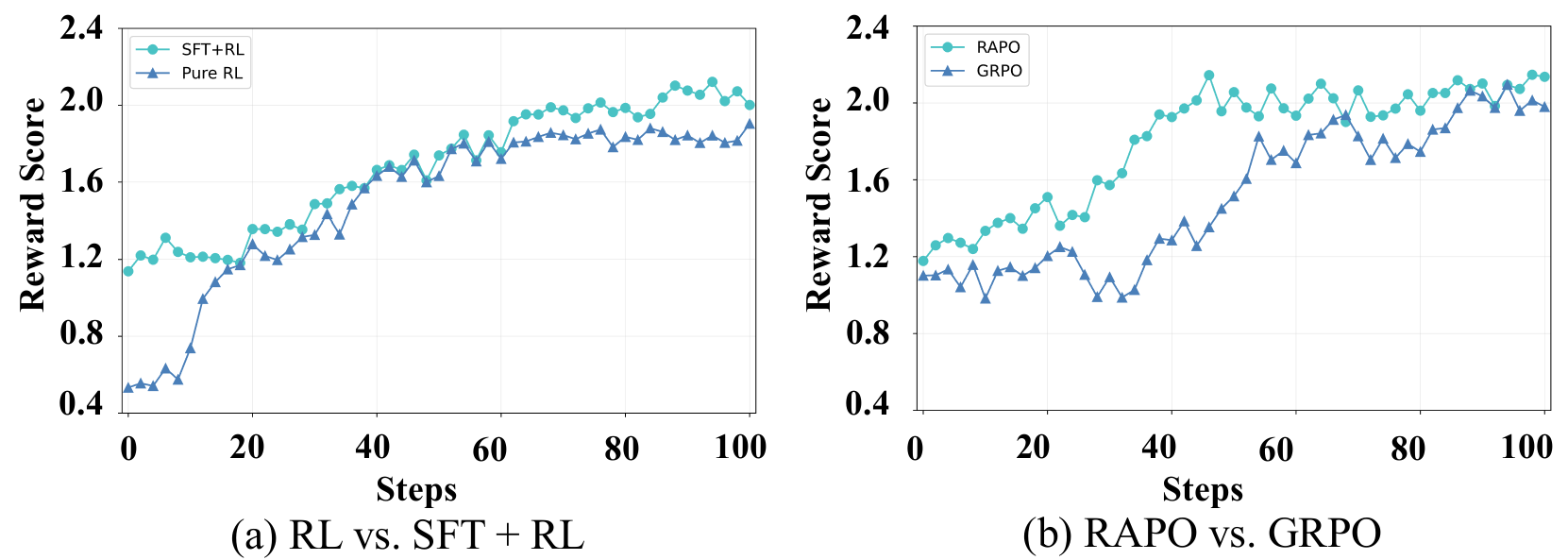}
\caption{Performance analysis of core components. (a) SFT initialization serves as a crucial warm-up, yielding higher initial rewards and accelerating convergence compared to Pure RL. (b) Our RAPO algorithm consistently outperforms the GRPO baseline, demonstrating superior sample efficiency and greater training stability throughout the optimization process.}
    \label{fig:reward}
\end{figure*}

\begin{figure*}[t]
    \centering
    \includegraphics[width=1\linewidth]{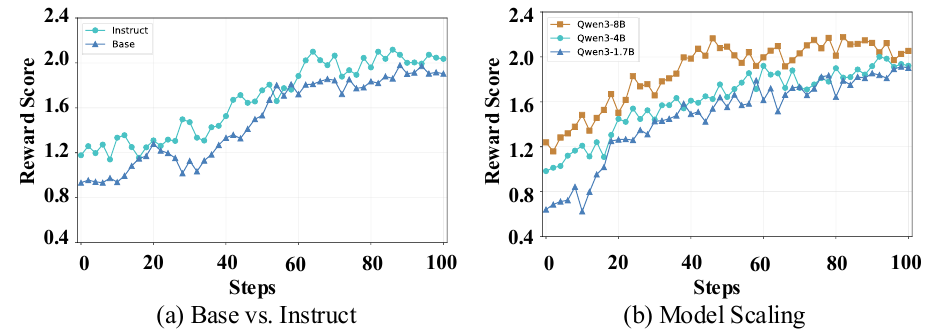}
\caption{Ablation study on model types and sizes. (a) The instruct-tuned model leverages superior initialization to maintain a consistent performance advantage over the base model throughout training. (b) Reward trajectories exhibit a clear positive correlation with model capacity, where larger parameter sizes consistently yield higher reward scores.}
    \label{fig:reward1}
\end{figure*}

\subsection{Ablation Studies of Training Strategies}
To systematically validate the design choices of our training framework, we conducted a series of ablation studies focusing on the critical components of the pipeline. Specifically, we investigate the impact of the SFT initialization, the necessity of the RFT stage, the effectiveness of the multi-objective reward shaping, and the advantages of our proposed RAPO algorithm over standard optimization baselines. The following subsections detail the experimental setup and results for each component, demonstrating their collective contribution to the final model performance.
\subsubsection{Effect of Supervised Fine-tuning.}
\begin{figure}[t]
    \centering
    \includegraphics[width=1\linewidth]{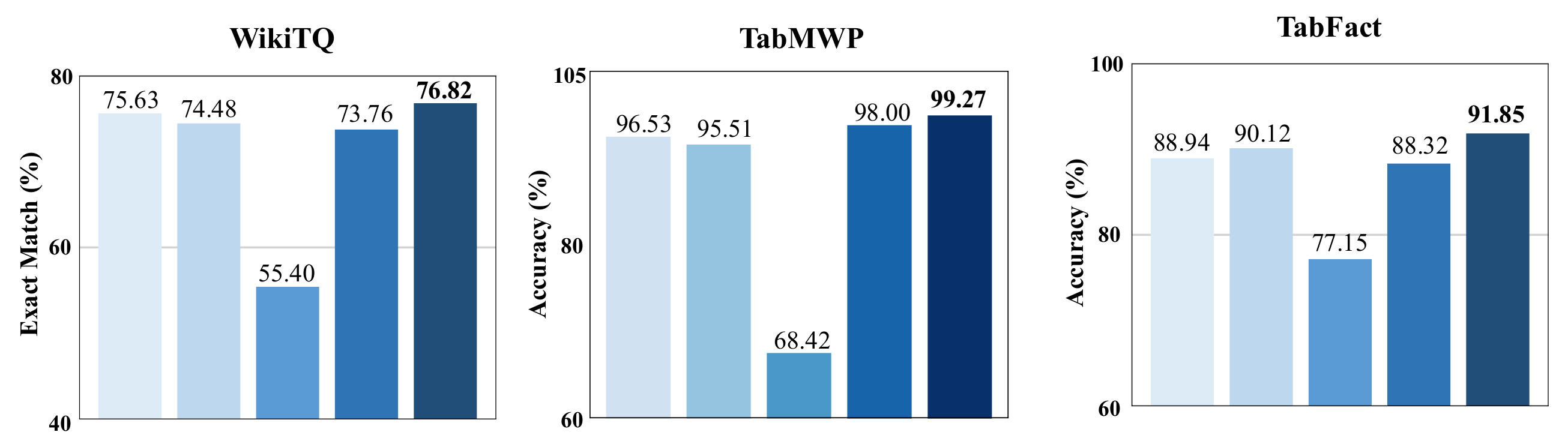}
    
    {\small
        \legenditem{color1}{w/o $R_{\text{tool}}$} \hspace{0.5em}
        \legenditem{color2}{w/o RAPO} \hspace{0.5em}
        \legenditem{color3}{w/o RFT} \hspace{0.5em}
        \legenditem{color4}{w/o SFT} \hspace{0.5em}
        \legenditem{color5}{Original}
    }
\caption{Ablation study on training strategies. We evaluate the contribution of each module by removing them individually from the full training framework. The results demonstrate that SFT, RFT, RAPO, and the auxiliary tool reward ($R_{\text{tool}}$) are all essential for achieving optimal performance, with the RFT stage proving most critical for accuracy.}
    \label{fig:Ablation}
\end{figure}

To determine the optimal strategy for the SFT stage, we conducted experiments on various training dataset sizes and epochs. The results are presented in Table~\ref{tab:sft_tradeoff}. They indicate that the optimal performance is achieved with a sample size of 200 at 1 epoch. Although we observe a slight regression in out-of-domain performance  compared to the base model, this trade-off is justifiable. The substantial gains in in-domain tasks provide a crucial warm-up for the model, establishing the necessary instruction-following capabilities and structural understanding. This robust initialization significantly accelerates convergence in the subsequent RL stage, making SFT an indispensable component despite the marginal OOD cost. Conversely, increasing the training to 2 epochs degrades results across the board, suggesting overfitting on our relatively small dataset.
As illustrated by the reward curve in Figure~\ref{fig:reward} (a), employing SFT as a warm-up phase enhances the efficiency of the subsequent RFT process. Compared to a pure RFT approach, the SFT-initialized model achieves a higher initial reward score during the early stage of training. As shown in Figure ~\ref{fig:Ablation}, the removal of the SFT phase leads to a degradation in model performance. This finding substantiates the necessity of SFT for our training methodology.

\subsubsection{Effect of Reinforcement Learning.}
In this study, we remove the RFT training stage and directly evaluate the model trained only with SFT. As shown in Figure~\ref{fig:Ablation}, this modification leads to a notable performance drop across benchmarks. On both WikiTQ and TabFact, the model’s accuracy decreases substantially, with similar declines observed across other evaluation metrics. These results indicate that while SFT provides a solid starting point by establishing baseline reasoning and code-generation skills, it does not equip the model with the adaptability and decision-making depth required for complex table reasoning tasks.
The RFT phase plays a crucial role in this regard by exposing the model to execution-based feedback, enabling it to refine tool-use policies and recover from intermediate errors.  Consequently, we identify RFT as a critical technique for boosting model performance, substantially enhancing generalization and enabling the model to maintain high reliability across diverse table reasoning scenarios.

\subsubsection{Effect of the Multi-objective Reward Design.}
We analyze the impact of our multi-objective reward function, with a particular focus on the contribution of our proposed Strategic Interaction Reward $R_{\text{tool}}$. To do so, we conducted an ablation study by removing this auxiliary component, training an agent that relies solely on the primary accuracy $R_{\text{acc}}$ and format rewards $R_{\text{format}}$. As shown in Figure~\ref{fig:Ablation}, removing the $R_{\text{tool}}$ term leads to a degradation in performance across all benchmarks.
Without the guidance of $R_{\text{tool}}$, the agent tended to generate less efficient and often erroneous reasoning chains. Specifically, the absence of the success incentive and efficiency penalty in $R_{\text{tool}}$ resulted in longer, more redundant tool-call sequences that were less likely to contribute to the correct final answer. The implicit curriculum provided by the reward's decay weight was also lost, leading to slower convergence.

\subsubsection{Effect of RAPO.}
Ablation studies were performed from two key perspectives to thoroughly assess the effectiveness of our proposed RAPO  algorithm. First, we compared the training dynamics of RAPO against the baseline GRPO algorithm. The reward score curves in Figure~\ref{fig:reward} (b) illustrate this comparison. Throughout the training process, the reward curve for RAPO is higher than that of GRPO. This indicates that RAPO not only converges faster but also exhibits a more stable training process with less volatility, demonstrating its advantages in improving the sample efficiency of reinforcement learning. Second, we quantified RAPO's contribution to the final task performance by removing it from the full model. As shown in Figure~\ref{fig:Ablation}, removing RAPO led to a degradation in performance on both benchmark datasets. The performance decreased on both the WikiTQ and TabFact datasets, showing a noticeable drop in Exact Match on WikiTQ and in Accuracy on TabFact.

\input{table/2-pure-sft}

\input{table/5-ablation}
\subsection{Ablation Studies of Inference Mechanisms}
To rigorously validate the individual contributions of our uncertainty-aware mechanisms, we conducted a comprehensive ablation study by systematically removing each module from the full TableMind++ framework. The results, summarized in Table~\ref{tab:ablation_mechanisms}, dissect the source of our performance gains and reveal the distinct and complementary roles of each component in ensuring robust reasoning.
First, the exclusion of memory-guided plan pruning precipitates the most severe performance degradation across all datasets, identifying it as the fundamental cornerstone of our framework. Without retrieving validated templates to mitigate epistemic uncertainty, the agent is prone to initiating fundamentally flawed reasoning trajectories, such as confusing aggregation logic with filtering operations. This confirms that accurate high-level planning is a strict prerequisite for success, as downstream execution tools are incapable of correcting structural flaws in the initial logic regardless of their robustness.
Second, removing confidence-based action refinement leads to a notable decline, particularly on benchmarks demanding precise code generation like TabMWP. This degradation underscores the module's vital role in dampening aleatoric noise. By intercepting low-confidence tokens and syntax errors before environmental interaction, it prevents minor generation artifacts, including hallucinated column names or malformed API calls, from cascading into fatal execution exceptions, thereby safeguarding the reliability of the execution phase.
Finally, disabling the uncertainty-weighted voting mechanism results in a consistent drop in accuracy across the board. Although the performance impact is more incremental compared to plan pruning, this component proves indispensable for stabilizing final outputs against stochastic variations. By synthesizing consensus from diverse paths and specifically penalizing low-confidence trajectories that might otherwise dominate a naive majority vote, it effectively filters out sporadic hallucinations. This confirms that confidence-aware ensemble consensus is crucial for achieving state-of-the-art performance.

\subsection{In-Depth Analysis}
To provide a holistic view of TableMind++, we conduct a multi-dimensional analysis beyond aggregate metrics. We first examine the impact of model initialization and scaling, then analyze the co-evolution of tool usage and execution reliability. Furthermore, we evaluate the inference efficiency trade-off and conclude with a fine-grained study on the shift in failure modes.
\subsubsection{Comparison of Model Variants}
Instruction tuning is hypothesized to align a foundational model’s general capabilities with the specific demands of following task-oriented commands, thereby providing a more favorable starting point for downstream optimization. We examine this hypothesis by comparing an Instruct-tuned model against a base model within our RFT framework. The reward curves in Figure~\ref{fig:reward1} (a) clearly validate this claim. The Instruct model not only starts with a substantial reward advantage—reflecting its prior adaptation to structured, goal-directed instructions—but also sustains this superiority throughout the entire training process, ultimately converging to a higher and more stable reward plateau. This behavior suggests that pre-aligning the model yields a more effective initial policy, reduces the exploration burden during RFT, and accelerates the acquisition of complex behaviors by enabling the agent to focus on fine-tuning execution strategies rather than learning fundamental task-following skills from scratch.

Beyond the impact of initialization paradigms, we further investigate the effect of model scaling to understand the relationship between model capacity and performance. We apply our full training pipeline to three Qwen3 models of increasing size: 1.7B, 4B, and 8B parameters. The results, shown in Figure~\ref{fig:reward1} (b), reveal a strong, positive correlation, consistent with the established scaling laws of LLMs. A clear performance hierarchy emerges from the very beginning of training and remains stable through convergence, with larger models consistently achieving higher rewards at every stage. This pattern suggests that greater representational capacity enables the model to capture more complex reasoning patterns, maintain richer intermediate state representations, and more effectively coordinate the multi-step processes involved in tool manipulation. These observations indicate that model scaling is not only a direct path to performance improvement but also a critical factor in mastering high-complexity reasoning tasks in this domain.

\begin{figure}[t]
    \centering
    \includegraphics[width=1\linewidth]{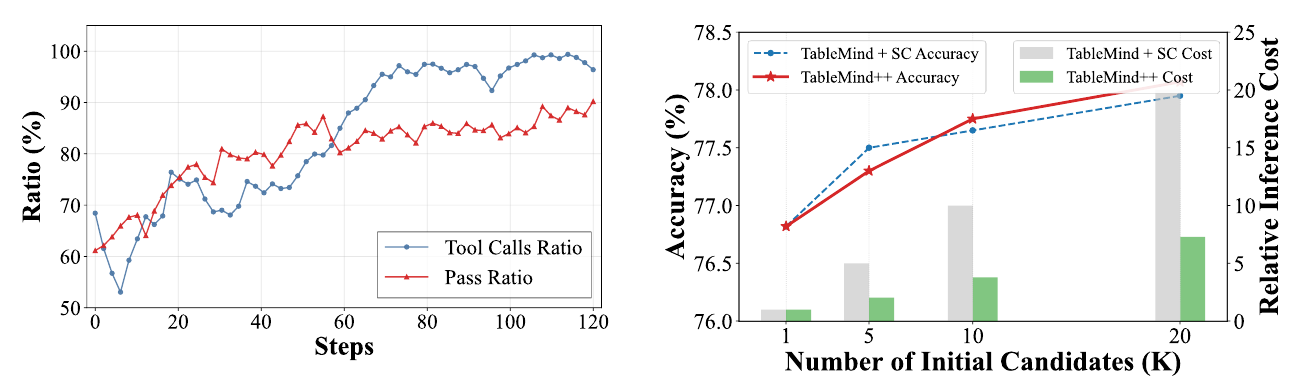}
\caption{Analysis of training dynamics and inference efficiency. The left panel displays the synchronized rise of the tool usage ratio and pass ratio indicating that the agent adopts tools as execution reliability improves. The right panel demonstrates that TableMind++ matches the accuracy of Self-Consistency with significantly lower computational cost.}

    \label{fig:efficiency_analysis}
\end{figure}
\subsubsection{Analysis of Tool Usage}
The evolution of the agent's policy for invoking tools is tracked by the blue line in the left panel of Figure~\ref{fig:efficiency_analysis}. The learning trajectory exhibits a clear and continuous progression with distinct phases. During the initial exploration phase, the agent's commitment to using tools is tentative and fluctuates as it balances direct textual generation with tool interaction, occasionally reverting to text-only outputs when tool use does not appear immediately beneficial. This is followed by a period of rapid discovery and exploitation, marked by a sharp upward inflection in the curve. Finally, as the training converges, the agent's strategy stabilizes with the tool-use ratio remaining near 100\%, indicating that the model has learned a near-deterministic policy that treats tool use as the optimal approach for the vast majority of tasks in this domain and applies it with high confidence.

Complementing this shift in usage policy, we further examine the pass ratio, illustrated by the red line in the same plot, which quantifies the reliability and correctness of the agent’s generated code. The curve shows a clear and sustained improvement in code-writing proficiency over the course of RFT training, indicating progressive refinement of the agent’s coding strategy. The model begins with a baseline pass ratio of approximately 60\%, reflecting the syntactic and semantic code generation capabilities developed during the SFT stage. During the RFT phase, the pass ratio steadily rises to nearly 90\%, with execution-based feedback guiding the agent to reinforce correct coding patterns, eliminate common sources of runtime errors, and align code more closely with task-specific requirements. These results suggest that the combination of SFT and RFT produces an agent that not only decides when to invoke a tool but also executes the invocation reliably and consistently across different problem settings.

\subsubsection{Analysis of Inference Efficiency} 
Building upon the robustly trained policy the right panel of Figure~\ref{fig:efficiency_analysis} evaluates the cost effectiveness during the inference phase. We compare the accuracy and cost trade off between TableMind++ and the standard Self Consistency baseline across varying numbers of candidate samples. The baseline method relies on brute force sampling to marginally improve accuracy which results in a prohibitive linear increase in computational cost as represented by the gray bars. In contrast TableMind++ achieves an accuracy comparable to the baseline at higher sample counts while maintaining a significantly lower inference cost as shown by the green bars. This substantial efficiency stems from our proposed inference time mechanisms including memory guided plan pruning and confidence based action refinement. By filtering out flawed reasoning paths early and correcting errors before execution our approach generates high quality solutions with fewer redundant attempts. Consequently TableMind++ effectively breaks the traditional dependency where higher accuracy inevitably requires exponentially higher computational expense proving its suitability for resource constrained real world deployment.

\begin{figure}[t]
    \centering
    \includegraphics[width=\linewidth]{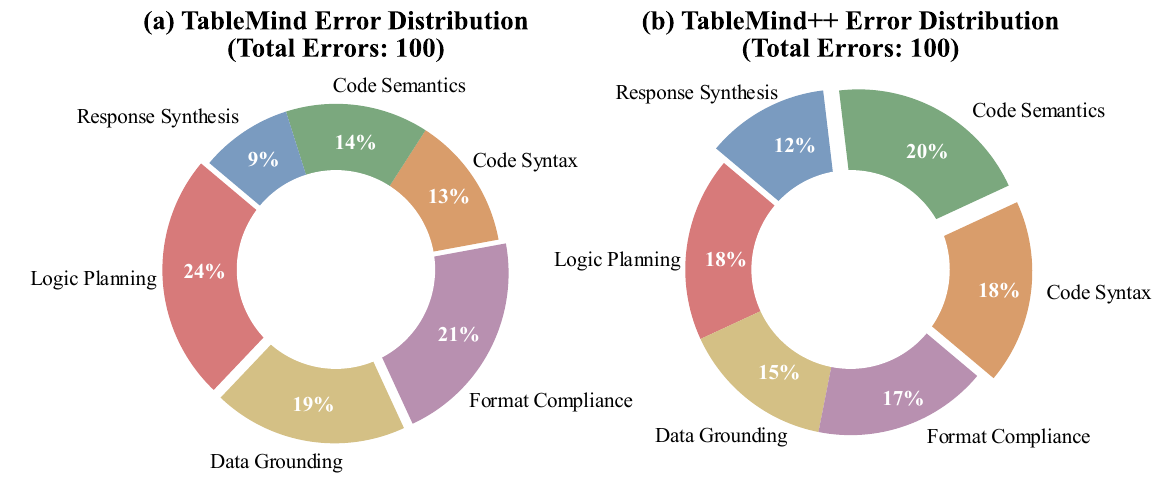}
\caption{Qualitative comparison of error distributions between TableMind (a) and TableMind++ (b). The results illustrate a fundamental shift in failure modes: TableMind++ effectively suppresses early-stage structural errors such as Logic Planning and Data Grounding, resulting in a cleaner error profile where remaining failures are concentrated in complex Code Semantics and Syntax, reflecting the intrinsic capability boundaries of the backbone model.}
    \label{fig:error_pie}
\end{figure}

\subsubsection{Fine-grained Error Analysis}
To gain deeper insights into the impact of our inference-time mechanisms, we conducted a manual error analysis on 100 randomly sampled failure cases from the Baseline TableMind and 100 from our proposed TableMind++. We categorized the errors into six distinct types spanning the agentic workflow: Logic Planning, Data Grounding, Format Compliance, Code Syntax, Code Semantics, and Response Synthesis. The comparative results, visualized in Figure~\ref{fig:error_pie}, unveil a significant Shift in Failure Modes. Initially, the Baseline model Figure~\ref{fig:error_pie}a is characterized by a high prevalence of foundational failures. Specifically, Logic Planning and Data Grounding constitute a major portion of the errors, indicating that the base agent frequently struggles with high-level strategy formulation and precise alignment between code literals and table content. In sharp contrast, TableMind++ Figure~\ref{fig:error_pie}b demonstrates a substantial suppression of these strategic and grounding faults. This qualitative improvement confirms that the Memory-Guided Plan Pruning mechanism effectively filters out flawed reasoning trajectories, while Action Refinement significantly mitigates hallucinations regarding column selection and value alignment.

With the effective mitigation of these low-level structural errors, the error landscape of TableMind++ undergoes a notable transformation, marking the emergence of complexity bottlenecks. We observe that the relative prominence of high-level execution errors particularly Code Semantics and Code Syntax increases noticeably. This phenomenon represents a relative inflation: as the agent becomes proficient at planning and grounding, it successfully reaches the deeper code execution stage more frequently, thereby exposing the intrinsic limitations of the backbone LLM in handling complex logic implementation and syntactic edge cases. Similarly, the persistent presence of Response Synthesis errors suggests that integrating complex execution results remains a challenging last-mile problem. In summary, TableMind++ improves reliability not merely by random optimization, but by systematically purifying the reasoning process eliminating avoidable structural errors and pushing the agent to its operational limits where only the most challenging semantic hurdles remain.

\input{table/3-Hyperparameter}
\input{table/6-Hyperparameter1}

\subsection{Hyperparameter Sensitivity Study}
We conducted a targeted study to identify the optimal settings for key hyperparameters during the training stage. This study focused on two core hyperparameters: the maximum number of interaction turns (Max Turns) and the generation Temperature. The results are summarized in Table~\ref{tab:hyperparam_training}. The results for Max Turns indicate that model performance significantly improves when the limit is relaxed from a restrictive setting to a moderate level. This suggests that a single turn is often insufficient for solving the complex, multi-step reasoning tasks inherent in these benchmarks. However, when the maximum turns are further increased beyond the optimal point, a slight degradation in performance is observed. This is likely because allowing excessive turns can increase the risk of error propagation or encourage the generation of redundant, unhelpful steps that interfere with the final outcome. Based on these findings, we identify the intermediate value as the optimal setting. Similarly, for the Temperature parameter, the results show a positive correlation between temperature and performance within the tested range. A higher temperature appears to encourage necessary diversity in exploration during the reinforcement learning process, whereas lower temperatures lead to overly deterministic responses that may get stuck in suboptimal policies. Consequently, the peak performance is achieved at the highest tested temperature.

Furthermore, we investigated the sensitivity of the inference-time mechanisms introduced in TableMind++, specifically the Memory Retrieval size ($K$) and the Confidence Threshold ($\tau$). The results are presented in Table~\ref{tab:hyperparam_inference}. For the Memory Retrieval size ($K$), while a moderate size generally offers the most robust performance across diverse benchmarks, we observe distinct task-dependent preferences. For simpler tasks like TabMWP, a smaller retrieval size yields marginally better results, likely because it minimizes the introduction of irrelevant noise from the memory bank. Conversely, for complex out-of-domain tasks such as FinQA, a larger retrieval size proves beneficial, suggesting that difficult reasoning requires a broader set of reference trajectories to cover the diverse solution space. Despite these variations, the intermediate value provides the best overall trade-off between context enrichment and noise control. Finally, regarding the Confidence Threshold ($\tau$) used for action refinement, the results exhibit an inverted U-shaped trend on most generative tasks, where a moderate threshold effectively balances error filtering against false rejections. However, for fact-verification tasks such as TabFact, a stricter threshold leads to superior performance. This indicates that tasks demanding high rigorousness benefit from more aggressive pruning of uncertain steps, whereas open-ended generation tasks require a more lenient threshold to preserve valid but slightly uncertain reasoning paths. Ultimately, we select the moderate settings as the default to ensure stability across all domains.

\begin{figure*}[t]
    \centering
    \includegraphics[width=1\linewidth]{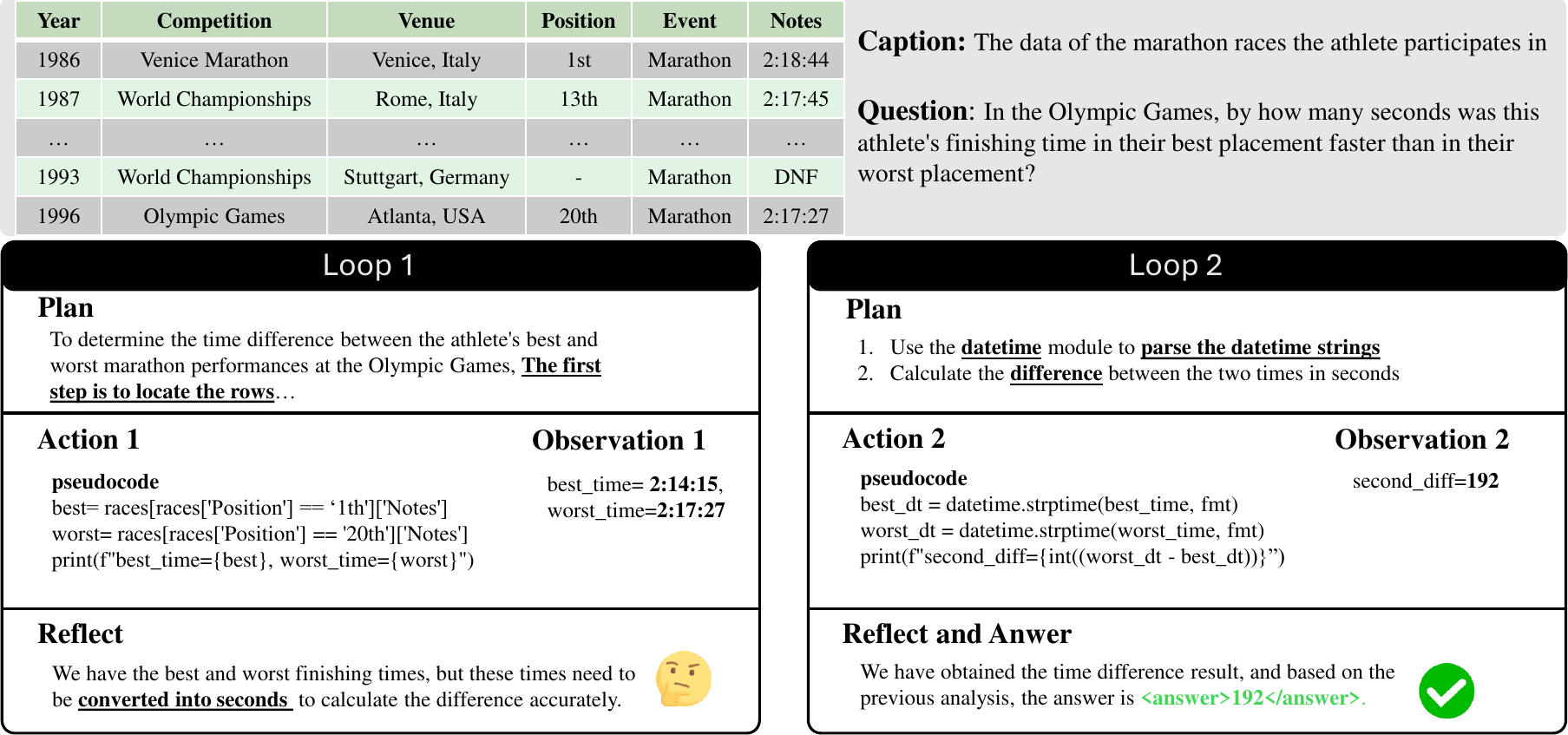}

\caption{A case study demonstrating the model's iterative plan-action-reflect loop. The agent initially retrieves raw data but identifies through self-reflection that the time strings require format conversion. This insight drives a refined plan in the second loop, showcasing its ability to dynamically decompose and solve complex queries.}

    \label{fig:case}
\end{figure*}

\subsection{Case Study}
We present a detailed case study in Figure~\ref{fig:case} that reveals the inner workings of TableMind++ on a representative WikiTQ problem. The model begins by formulating a simple initial plan, successfully extracting two time strings from the table. Crucially, instead of proceeding with a naïve computation that would yield an incorrect result, the model enters a reflection state. In this stage, it identifies that a simple subtraction is infeasible because the time data is stored as textual strings rather than numerical values. This self-reflection prevents premature errors and guides the model toward a more robust solution strategy. Armed with this insight, the model devises a refined plan: it first converts the textual strings into a proper time format suitable for arithmetic operations, and then computes the precise difference. To implement this plan, the model generates executable code using Python's datetime module. The code runs flawlessly, producing the correct numerical output of 192 seconds. Finally, the model synthesizes this raw numerical result into a well-formatted answer, ready to be returned. This example vividly illustrates that TableMind++ is capable of decomposing complex queries, reflecting on intermediate steps, self-correcting its reasoning strategy, and strategically leveraging external tools to solve table-based problems.

%% file: table/4-dataset.tex
\begin{table}[htbp]
	\centering
\caption{Dataset statistics for training and evaluation. In-domain datasets (TabFact, TabMWP, WTQ) are used for training, where Test (RL) denotes the subset reserved for monitoring performance during the RL training phase. Out-of-domain datasets (HiTab, FinQA) are utilized exclusively for final evaluation.}
	\label{tab:dataset_stats}
	\begin{tabular}{lccccc}
		\toprule
		\textbf{Split} & \textbf{TabFact} & \textbf{TabMWP} & \textbf{WTQ} & \textbf{HiTab} & \textbf{FinQA} \\
		\midrule
		Training   & 3,500 & 3,500 & 1,000 & -- & -- \\
		Test (RL)  & 800  & 800 & 800 & -- & -- \\
		Evaluation & 12,779  & 7,686 & 4,344 & 1,349 & 1,138 \\
		\bottomrule
	\end{tabular}
\end{table}

%% file: table/1-main-result.tex
\begin{table*}[t]
  \centering
\caption{Main results (\%) of different baselines over the benchmarks. In each column, the best result is marked in \textbf{bold} and the second-best result is \underline{underlined}. Our proposed TableMind++ consistently achieves the best performance across both in-domain and out-of-domain tasks, demonstrating exceptional capability in handling complex structural and numerical reasoning.}
  \setlength{\tabcolsep}{8pt}
    \begin{tabular}{llccc >{\centering\arraybackslash}p{1.2cm} >{\centering\arraybackslash}p{1.2cm}}
    \toprule
    \multirow{2}{*}{} & \multirow{2}{*}{Models} & \multicolumn{3}{c}{In-domain Performance} & \multicolumn{2}{c}{Out-of-domain Performance} \\
    \cmidrule(lr){3-5} \cmidrule(lr){6-7}
          &       & Wikitq & TabMWP & TabFact & HiTab & FinQA \\
    \midrule
\multirow{4}{*}{Proprietary} & GPT-4.1 & 59.13 & 90.34 & 83.94 & 61.42 & 6.45 \\
          & GPT-5 & 77.42 & 96.12 & 90.05 & 44.52 & 28.93 \\
          & Gemini-2.0-flash & 72.54 & 91.56 & 81.03 & 76.08 & 19.62 \\
          & Gemini-2.5-flash & 75.88 & 95.89 & 89.65 & 52.79 & 33.45 \\
    \midrule
    \multirow{3}{*}{Open-source} &
    Qwen3-8B & 50.47 & 60.86 & 70.34 & 63.12 & 27.83 \\
    & Qwen2.5-72B-Instruct & 68.56 & 97.45 & 87.34 & 68.02 & 35.83 \\
          & Deepseek-R1 & 74.63 & 98.03 & 86.25 & 70.37 & 37.42 \\
    \midrule
    \multirow{3}{*}{Training-free} & Tab-CoT & 60.43 & 92.21 & 79.05 & 60.64 & 27.81 \\
          & PoTable & 69.56 & 95.76 & 88.93 & 61.96 & 24.93 \\
          & Chain-of-Table & 68.31 & 96.17 & 87.78 & 64.76 & 29.47 \\
    \midrule
    \multirow{5}{*}{Tuning-based} & TableLlama-7B & 53.32 & 83.88 & 82.53 & 58.73 & 15.73 \\
          & TableGPT2-7B & 62.46 & 80.43 & 78.87 & 65.42 & 17.69 \\
          & Table-R1 & 74.86 & 96.02 & 87.17 & 69.24 & 41.27 \\
          & TableMind & \underline{76.82} & \underline{99.27} & \underline{91.85} & \underline{71.95} & \underline{42.02} \\
    \cmidrule{2-7}          
          & TableMind++ & \textbf{78.07} & \textbf{99.57} & \textbf{93.73} & \textbf{73.69} & \textbf{45.48} \\
    \bottomrule
    \end{tabular}%
  \label{tab:main result}%
\end{table*}  

%% file: table/2-pure-sft.tex
\begin{table}[t]
  \centering
\caption{Performance trade-off during the SFT stage. The selected configuration (Epoch 1, 200 samples) achieves substantial in-domain gains with minimal loss in OOD generalization. Crucially, this serves as an effective warm-up phase, facilitating faster convergence and better stability in the downstream reinforcement learning process.}
  
  \setlength{\tabcolsep}{4pt} 
    \begin{tabular}{cc ccc >{\centering\arraybackslash}p{1.2cm} >{\centering\arraybackslash}p{1.2cm}}
    \toprule
    \multirow{2}{*}{Epoch} & \multirow{2}{*}{Sample Size} & \multicolumn{3}{c}{In-domain Performance} & \multicolumn{2}{c}{Out-of-domain Performance} \\
    \cmidrule(lr){3-5} \cmidrule(lr){6-7}
          &       & Wikitq & TabMWP & TabFact & HiTab & FinQA \\
    \midrule
    0     & -     & 50.47 & 60.86 & 70.34 & \textbf{63.12} & \textbf{27.83 }\\
    \midrule
    \multirow{3}[2]{*}{1} & 100   & 52.52 & 62.49 & 70.16 & 62.85 & 27.41 \\
          & 150   & 53.65 & 62.63 & 71.32 & 62.40 & 27.15 \\
          & 200   & \textbf{55.40} & \textbf{68.42} & \textbf{77.15} & 62.15 & 26.88 \\
    \midrule
    \multirow{3}[2]{*}{2} & 100   & 50.12 & 60.62 & 70.12 & 61.50 & 26.05 \\
          & 150   & 50.42 & 61.48 & 70.29 & 60.88 & 25.42 \\
          & 200   & 51.69 & 62.28 & 73.62 & 59.74 & 24.19 \\
    \bottomrule
    \end{tabular}%
  \label{tab:sft_tradeoff}%
\end{table}%

%% file: table/5-ablation.tex
\begin{table}[t]
  \centering
  \caption{Ablation study on inference-time mechanisms. We evaluate the contribution of each module by removing them individually from the full TableMind++ framework. The results demonstrate that memory-guided plan pruning, confidence-based action refinement, and uncertainty-weighted voting are all essential for achieving optimal performance.}
  
  \setlength{\tabcolsep}{4pt} 
  
  \begin{tabular}{l ccc >{\centering\arraybackslash}p{1.2cm} >{\centering\arraybackslash}p{1.2cm}}
    \toprule
    \multirow{2}{*}{Model / Variant} & \multicolumn{3}{c}{In-domain Performance} & \multicolumn{2}{c}{Out-of-domain Performance} \\
    \cmidrule(lr){2-4} \cmidrule(lr){5-6}
          & Wikitq & TabMWP & TabFact & HiTab & FinQA \\
    \midrule
    \textbf{TableMind++ (Full)} & \textbf{78.07} & \textbf{99.57} & \textbf{93.73} & \textbf{73.69} & \textbf{45.48} \\
    \midrule
    w/o Weighted Voting & 77.85 & 99.45 & 93.25 & 73.15 & 44.85 \\
    w/o Action Refinement & 77.42 & 99.30 & 92.80 & 72.50 & 43.60 \\
    w/o Plan Pruning      & 77.10 & 99.15 & 92.45 & 72.10 & 42.95 \\
    \midrule
    TableMind (Base)      & 76.82 & 99.27 & 91.85 & 71.95 & 42.02 \\
    \bottomrule
    \end{tabular}%
  \label{tab:ablation_mechanisms}%
\end{table}

%% file: table/3-Hyperparameter.tex
\begin{table}[t]
  \centering
\caption{Hyperparameter sensitivity analysis regarding the Training (RFT) stage. We systematically evaluate the impact of the maximum number of tool-call turns and the rollout temperature on model performance. The results highlight that setting the turn limit to 3 and the temperature to 1.0 achieves the optimal balance between reasoning depth and exploration efficiency.}
  \setlength{\tabcolsep}{4pt}
  
  \begin{tabular}{cc ccc >{\centering\arraybackslash}p{1.2cm} >{\centering\arraybackslash}p{1.2cm}}
    \toprule
    \multirow{2}{*}{Hyperparameter} & \multirow{2}{*}{Value} & \multicolumn{3}{c}{In-domain Performance} & \multicolumn{2}{c}{Out-of-domain Performance} \\
    \cmidrule(lr){3-5} \cmidrule(lr){6-7}
          &       & Wikitq & TabMWP & TabFact & HiTab & FinQA \\
    \midrule
    \multirow{3}[2]{*}{Max Turns} & 1     & 74.35 & 97.23 & 89.83 & 68.52 & 38.15 \\
          & 3     & \textbf{76.82} & \textbf{99.03} & \textbf{92.33} & \textbf{71.95} & \textbf{42.02} \\
          & 5     & 75.65 & 98.56 & 90.55 & 70.88 & 40.65 \\
    \midrule
    \multirow{3}[2]{*}{Temperature} & 0.6   & 73.35 & 96.58 & 88.43 & 68.12 & 38.45 \\
          & 0.8   & 75.51 & 98.84 & 91.25 & 70.25 & 40.12 \\
          & 1.0     & \textbf{76.82} & \textbf{99.03} & \textbf{92.33} & \textbf{71.95} & \textbf{42.02} \\
    \bottomrule
    \end{tabular}%
  \label{tab:hyperparam_training}%
\end{table}%

%% file: table/6-Hyperparameter1.tex
\begin{table}[t]
  \centering
  \caption{Sensitivity analysis for Inference-time mechanisms in TableMind++. This table investigates the impact of Memory Retrieval size ($K$) and the Confidence Threshold ($\tau$). While $K=5$ and $\tau=0.8$ are selected as the default settings for their robust overall performance, some specific tasks benefit from alternative configurations.}
  
  \setlength{\tabcolsep}{4pt}
  
  \begin{tabular}{cc ccc >{\centering\arraybackslash}p{1.2cm} >{\centering\arraybackslash}p{1.2cm}}
    \toprule
    \multirow{2}{*}{Hyperparameter} & \multirow{2}{*}{Value} & \multicolumn{3}{c}{In-domain Performance} & \multicolumn{2}{c}{Out-of-domain Performance} \\
    \cmidrule(lr){3-5} \cmidrule(lr){6-7}
          &       & Wikitq & TabMWP & TabFact & HiTab & FinQA \\
    \midrule
    \multirow{3}[2]{*}{Memory $K$} & 3     & 77.20 & \textbf{99.65} & 92.50 & 72.10 & 44.20 \\
          & 5     & \textbf{78.07} & 99.57 & \textbf{93.73} & \textbf{73.69} & 45.48 \\
          & 10    & 77.55 & 99.35 & 93.10 & 72.85 & \textbf{45.82} \\
    \midrule
    \multirow{3}[2]{*}{Conf. Threshold $\tau$} & 0.6   & 76.90 & 98.80 & 91.90 & 71.50 & 43.80 \\
          & 0.8   & \textbf{78.07} & \textbf{99.57} & 93.73 & \textbf{73.69} & \textbf{45.48} \\
          & 0.9   & 77.80 & 99.20 & \textbf{93.88} & 73.10 & 45.10 \\
    \bottomrule
    \end{tabular}%
  \label{tab:hyperparam_inference}%
\end{table}

%% file: 6-Conclusion.tex
\section{Conclusion}
In this paper, we present TableMind++, an uncertainty-aware programmatic agent designed to perform rigorous  table reasoning. By combining a two-stage training foundation of Supervised Fine-tuning and Rank-Aware Policy Optimization (RAPO) with a novel inference architecture, our framework effectively mitigates reasoning risks. The integration of Memory-Guided Plan Pruning filters epistemic hallucinations, while Confidence-Based Action Refinement corrects aleatoric execution noise. Furthermore, Dual-Weighted Trajectory Aggregation synthesizes these outputs to ensure the final consensus is well-calibrated. Extensive experiments across multiple benchmarks demonstrate that TableMind++ achieves state-of-the-art performance, significantly outperforming competitive baselines. Future work will focus on extending this paradigm to multi-table environments, integrating a broader spectrum of data science tools, and further optimizing the sample efficiency of the reinforcement learning process.
